\documentclass[runningheads]{llncs}

\usepackage[final]{eccv}

\usepackage{eccvabbrv}
\usepackage[algo2e]{algorithm2e} 
\usepackage{amsmath,amssymb,bm}
\SetKwInput{KwHyper}{HyperParameter}
\SetKwInput{KwInput}{Input (Optional)}
\SetKwInput{KwOutput}{Output}
\usepackage{booktabs}
\usepackage{multirow}
\usepackage{algorithm}
\usepackage{algorithmic}
\usepackage{graphicx}
\usepackage{booktabs}
\usepackage{multirow}
\usepackage{wrapfig}
\usepackage{graphicx}
\usepackage{booktabs}

\usepackage[accsupp]{axessibility}  

\usepackage{hyperref}

\usepackage{orcidlink}

\begin{document}

\title{SemConFlow: Semantic Grounding of Holistic Co-Speech Gesture Generation with Contrastive Flow-Matching}

\titlerunning{SemConFlow}

\author{Lanmiao Liu\inst{1,2,3}\orcidlink{0000-0002-2384-0657} \and
Esam Ghaleb\inst{1,2}\orcidlink{0000-0002-0603-9817} \and
Asl\i~Özy\"{u}rek\inst{1,2}\orcidlink{0000-0003-4914-2643} \and Zerrin Yumak\inst{3}\orcidlink{0000-0002-0028-5806} }


\authorrunning{Liu et al.}

\institute{Max Planck Institute for Psycholinguistics \and
Donders Institute for Brain Cognition and Behaviour
\and
Utrecht University\\
\email{\{lanmiao.liu, esam.ghaleb, asli.ozyurek\}@mpi.nl, z.yumak@uu.nl} } 

\maketitle

\begin{abstract}
    While the field of co-speech gesture generation has seen significant advances, producing holistic, semantically grounded gestures remains a challenge. Existing approaches rely on external semantic retrieval methods, which limit their generalisation capability due to dependency on predefined linguistic rules. Flow-matching-based methods produce promising results; however, the network is optimised using only semantically congruent samples without exposure to negative examples, leading to learning rhythmic gestures rather than sparse motion, such as iconic and metaphoric gestures. Furthermore, by modelling body parts in isolation, the majority of methods fail to maintain cross-modal consistency.  We introduce a Contrastive Flow Matching-based co-speech gesture generation model that uses mismatched audio–text conditions as negatives, training the velocity field to follow the correct motion trajectory while repelling semantically incongruent trajectories. Our model ensures cross-modal coherence by embedding text, audio, and holistic motion into a composite latent space via cosine and contrastive objectives. Extensive experiments and a user study demonstrate that our proposed approach outperforms state-of-the-art methods on two datasets, BEAT2 and SHOW. Our project page is available at: \href{https://marcos452.github.io/HoliticSemGes/}{https://marcos452.github.io/HoliticSemGes/}
\end{abstract}

\section{Introduction}
\label{sec:intro}
Human communication is multimodal: speech is produced and interpreted together with coordinated bodily visual signals, such as co-speech gestures, as part of a single composite message \cite{Holler2019}. Importantly, combinations of visual cues (e.g., head, gestures, and facial signals) can change perceived communicative intent relative to each cue in isolation, indicating a compositional, cross-channel organisation of meaning \cite{Trujillo2024}. Humans also integrate modalities by \emph{congruency}: content-aligned speech–gesture signals facilitate comprehension, whereas incongruent pairings can disrupt or mislead interpretation \cite{Kelly2010}. For generative modelling, this raises the bar beyond motion realism: generated motion must be semantically congruent, i.e., grounded in and aligned with the speech, and globally consistent across articulators (i.e., body parts).

Despite recent progress in holistic co-speech motion generation~\cite{zhang2025semtalk,liu2025gesturelsm,chen2024diffsheg,zhu2023taming,liu2025intentional,zhang2025kinmo,paar2026duogesture}, two challenges remain. First, most holistic pipelines still perform \emph{multimodal alignment and conditioning in a part-wise manner}: motion is tokenised into body-part latents, and speech and semantics embeddings are projected and injected separately for each region rather than being anchored to a single composite full-body representation. This factorised alignment limits cross-region coordination, so the model can satisfy semantic cues in one channel while others drift, leading to cross-articulator inconsistency (e.g., semantically informative hand motion paired with neutral or mismatched head and face behaviour).
\begin{figure*}[t!]
  \centering
  \includegraphics[width=0.99\linewidth]{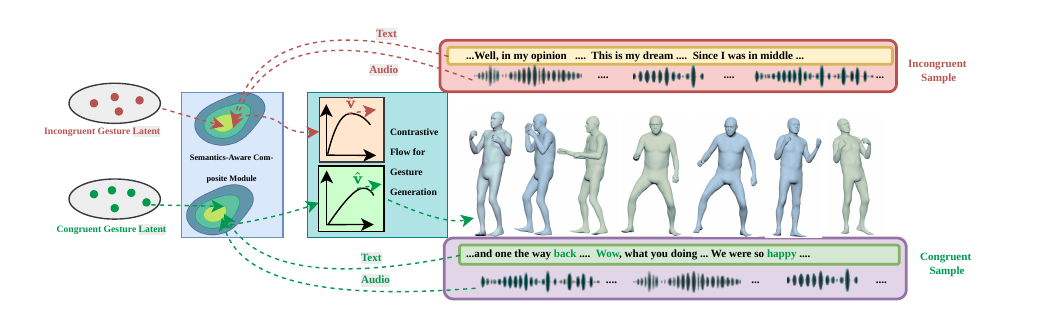}
  \caption{Overall idea of our model. Audio and text transcripts are encoded into a shared semantic representation, where congruent (matching) and incongruent (mismatching) speech and gesture latents form contrastive pairs. Contrastive flow matching learns motion trajectories that align with the correct semantic context while diverging from mismatched samples, producing coherent full-body gestures.}
  \label{fig:1}
\end{figure*}
\emph{Second, and most importantly,} the dominant generative paradigms such as Diffusion Models \cite{chen2024diffsheg, zhu2023taming, alexanderson2023listen,ning2024dctdiff} and Flow Matching \cite{liu2025gesturelsm,liu2023human} optimise for distribution matching via distance-based (e.g., mean square error) or likelihood-style objectives. These objectives encourage motions that look realistic at first sight, but they lack the semantic grounding that links the motion with the \emph{meaning} of the spoken content. Moreover, they lack explicit comparison against semantically \emph{incongruent} alternatives, i.e., negative samples. Hence, they optimise to generate generic, beat-like gestures aligned with the rhythm of the speech, rather than sparse motion, such as iconic and metaphoric gestures that carry the intended semantics.

To address these two challenges, we present HolisticSemGes, an approach for high-fidelity holistic co-speech gesture generation with two complementary components.
First, we introduce the \textbf{Semantics-Aware Composite Module (SACM)}. SACM explicitly shapes motion latents to be congruent with speech semantics by aligning speech-text embeddings with pre-quantised motion latents at the frame level. It uses a contrastive structure pulling matched speech--motion pairs together, while pushing apart mismatched (semantically incongruent) pairs, whereas prior methods such as \cite{zhang2025semtalk, liu2024emage} neglect such incongruent relationships.
In contrast with previous holistic work that encodes body parts separately \cite{liu2024emage}, we treat them as a composite full-body representation to preserve the cross-channel organisation of meaning.
Second, we propose a \textbf{Contrastive Flow Matching (CFM)} \cite{stoica2025contrastive}- based co-speech gesture generation framework, different from previous work based on standard flow matching \cite{liu2025gesturelsm}. While standard flow matching \cite{dao2023flow} learns a conditional velocity field that transports noise to the target motion, it does not explicitly teach the model which motions are \emph{semantically wrong} for a given utterance. CFM introduces mismatched audio--text conditions as \emph{negative} context and trains the flow to align with the ground-truth trajectory as well as diverge from trajectories induced by incongruent semantics. This contrastive constraint makes the generation path from noise to motion more directionally stable and improves semantic grounding.  Figure~\ref{fig:1} presents the general idea of our approach. To summarise, our contributions are as follows:
\begin{itemize}
    \item We introduce the \emph{Semantics-Aware Composite Module} that learns a shared latent space across audio, text, and motion representations. This module improves cross-modal consistency by embedding the input modalities into a semantically aligned composite latent space. 
    \item We propose a \emph{Contrastive Flow Matching} based co-speech gesture generation framework, encouraging the latent flow to not only follow the correct motion trajectory but also avoid incorrect or mismatched speech-gesture directions. This leads to stable training, improved gesture diversity, and enhanced speech–gesture alignment.
    \item On two benchmark datasets, BEAT2 \cite{liu2024emage} and SHOW \cite{yi2023generating}, we provide extensive objective and subjective evaluation and ablation studies showing state-of-the-art performances across several metrics. 
\end{itemize}

\section{Related Work}
\label{sec:realtedwork}
\subsection{Holistic Co-Speech Gesture Generation}
Early methods addressed individual body regions in isolation and relied on skeletal representations \cite{ginosar2019learning,yoon2020speech,kucherenko2020gesticulator,liu2025contextual}, which inherently prevent holistic semantic coherence. Recent work \cite{yi2023generating,liu2024emage, chen2024diffsheg, liu2025gesturelsm,guo2025fasttalker} have focused on holistic generation, including a combination of visual cues using mesh-based representations such as SMPL-X \cite{pavlakos2019expressive}. TalkSHOW \cite{yi2023generating} and EMAGE \cite{liu2024emage} train body part-specific VQ-VAEs \cite{van2017neural} with autoregressive decoding, improving temporal coherence, yet learned codebooks are optimised for reconstruction fidelity alone with no semantic grounding. Diffusion-based methods such as DiffSHEG \cite{chen2024diffsheg} achieve high sample diversity via audio-conditioned reverse diffusion, but require hundreds of inference steps and provide no mechanism for semantic guidance in the generation. Existing holistic approaches improve full-body realism \cite{liu2025gesturelsm,guo2025fasttalker}, but they lack semantic grounding that explicitly aligns motion trajectories with the speech content.

\subsection{Semantic Gesture Generation}
Methods that focus on semantic grounding are mostly skeleton-based approaches and cannot generate holistic motion \cite{yoon2020speech,liang2022seeg,zhi2023livelyspeaker,liu2025semges}. GestureDiffuCLIP \cite{ao2023gesturediffuclip} includes whole body and head movements in a holistic manner and constructs a shared audio–gesture embedding space via CLIP-inspired \cite{radford2021learning} contrastive pretraining. A key limitation of this approach is the decoupling of the alignment objective from the generative model. Contrastive pretraining shapes the conditioning space, but the generative objective reconstruction loss remains semantically agnostic, leaving semantically misaligned outputs unpunished during training. Semantic Gesticulator \cite{zhang2024semantic} and RAG-Gesture \cite{mughal2025retrieving} address semantic grounding; however, they require an extra retrieval stage with predefined semantic guidance using linguistic insights. SemTalk \cite{zhang2025semtalk} generates both rhythm and semantics-aware gestures; however, they are combined in a separate stage, leading to less realistic and diverse motion. In contrast, our method does not rely on an external retrieval stage or a separate pipeline for rhythm and semantics. Instead, it learns semantic grounding in an end-to-end manner with speech-text embeddings and semantics-guided contrastive flow-matching-based generation.

\subsection{Generative Methods for Holistic Gesture Generation}
Autoregressive Transformer decoders combined with (R)VQVAE discretisations are often used as the generative method \cite{yi2023generating,zhang2025semtalk,liu2024emage,ning2026spectrum}; however, they suffer from compounding errors over long sequences. Diffusion models~\cite{chen2023executing,zhu2023taming} achieve high fidelity and diversity, yet they require hundreds of denoising steps. Flow matching \cite{liu2025gesturelsm} learns a deterministic velocity field, enabling high-quality generation in a few steps more efficiently and with higher stability. MotionFlow~\cite{HuMotionFM2024} demonstrates its advantages for full-body motion synthesis. GestureLSM\cite{liu2025gesturelsm} introduces a flow matching-based co-speech gesture generation method; however, the network does not learn to punish semantically ungrounded motions. Our approach addresses this limitation using Contrastive Flow Matching (CFM), which embeds cosine similarity-based contrastive alignment into velocity field learning. This ensures that the deterministic path remains semantically faithful to the speech content throughout generation rather than converging to a statistically plausible but semantically averaged output.

\section{Methodology}
\label{sec:Methodology}
We propose HolisticSemGes, a novel approach for multimodal-driven holistic co-speech gesture generation over full-body SMPLX parameters. Sec \ref{sec:motion} presents representation learning per-body-part Residual VQ-VAEs that compress full-body SMPL-X motion into compact latent sequences. Sec \ref{sec:semantic} introduce how to generate semantically coherent gestures, conditioned on speech acoustics and lexical content, with contrastive flow matching.

\subsection{Problem Formulation}
Our objective is to generate semantically grounded holistic gestures by jointly modelling multiple body parts conditioned on speech audio and textual semantics.
We represent the gesture sequence for each part $r \in \mathcal{R}$, where 
$\mathcal{R} = \{\text{hand}, \text{upper}, \text{lower}, \text{face}\}$ as 
$\mathbf{G}^{r} = \{\mathbf{g}_i^{r}\}_{i=1}^{L} \in \mathbb{R}^{L \times J}$, 
where $L$ denotes the number of time steps (i.e., sequence length) and $J$ the number of joints within the corresponding region. To capture realistic motion dynamics, we first learn a motion generator $\mathcal{M}_g$ under a motion-aware representation learning framework. The learning objective of $\mathcal{M}_g$ is defined as follows:
\begin{align}
\arg \min_{\mathcal{M}_g} \left\| \mathbf{G}^{r} - \mathcal{M}_g(\mathbf{G}^{r}) \right\|
\end{align}
Next, each motion stream undergoes encoding through a motion encoder $\mathcal{E}_m$, transforming kinematic data into intermediate feature representations $\mathbf{Z}^g$. Given multimodal inputs consisting of raw speech audio $\mathbf{A} = \{\mathbf{a}_i\}_{i=1}^{L}$, text-based semantic embeddings $\mathbf{S} = \{\mathbf{s}_i\}_{i=1}^{L}$, as well as mismatched (incongruent) speech audio 
$\mathbf{\hat{A}} = \{\mathbf{\hat{a}}_i\}_{i=1}^{L}$ and mismatched semantic embeddings $\mathbf{\hat{S}} = \{\mathbf{\hat{s}}_i\}_{i=1}^{L}$, our goal is to learn a multimodal generator $\mathcal{M}_{m}$ that maps $(\mathbf{A}, \mathbf{S}, \mathbf{\hat{A}}, \mathbf{\hat{S}})$ to a latent representation. The overall objective is defined as follows:
\begin{equation}
\arg \min_{\mathcal{M}_{m}}
\left\|
\mathbf{Z}^g
-
\mathcal{M}_{m}(\mathbf{A}, \mathbf{\hat{A}},\mathbf{S},\mathbf{\hat{S}})
\right\|
\end{equation}

\subsection{1st Stage: Motion Aware Representation Learning}
\label{sec:motion}
We start with a motion prior learning stage that establishes fundamental movement patterns through the utilization of hierarchical RVQ-VAE with temporal compression inspired by \cite{guo2024momask,liu2024emage,zhang2025semtalk}.  This multi-scale tokenization approach uses residual vector quantizers (RVQ)\cite{lee2022autoregressive,borsos2023audiolm,zeghidour2021soundstream}  to discretize continuous gesture motions of hand, upper body, lower body and facial expressions into structured token sequences. 

Given body-part-specific motion streams $\mathbf{G}^{r}$ for each body part $r$, 
we learn motion encoder $\mathcal{E}^{r}_m$ to extract latent representations. 
The encoded features are then quantised using region-specific codebooks $\mathbf{Q}^{r}$, and subsequently decoded by the motion decoder $\mathcal{D}^{r}_m$ to reconstruct the gesture sequence $\mathbf{\hat{G}}^{r}$ for each body part. The learning process optimises a composite objective function that balances reconstruction fidelity with quantisation stability across all body regions. We formulate the loss as:

\begin{align}
\mathcal{L}_{total} &= \mathcal{L}_{rec}(\mathbf{G}^{r},\mathbf{\hat{G}}^{r}) + \mathcal{L}_{com}(\mathrm{sg}[\mathbf{G}^{r}],\mathbf{Q}^r(\mathbf{Z}^{g}))
\end{align}
where the first term denotes reconstruction loss between the generated gesture and the ground truth, the second term implements the VQ-VAE commitment loss\cite{van2017neural}.

\subsection{2nd Stage: Generating Semantically Coherent and Grounded Gestures}
\label{sec:semantic}
The objective here is to condition motion generation on speech acoustics and text semantics, achieving meaningful, holistic gesture generation. We therefore apply efficient and robust mechanisms to map speech and text content to the latent motion space, learned in the first stage.

\begin{figure*}[htbp]
  \centering
  \includegraphics[width=0.99\linewidth]{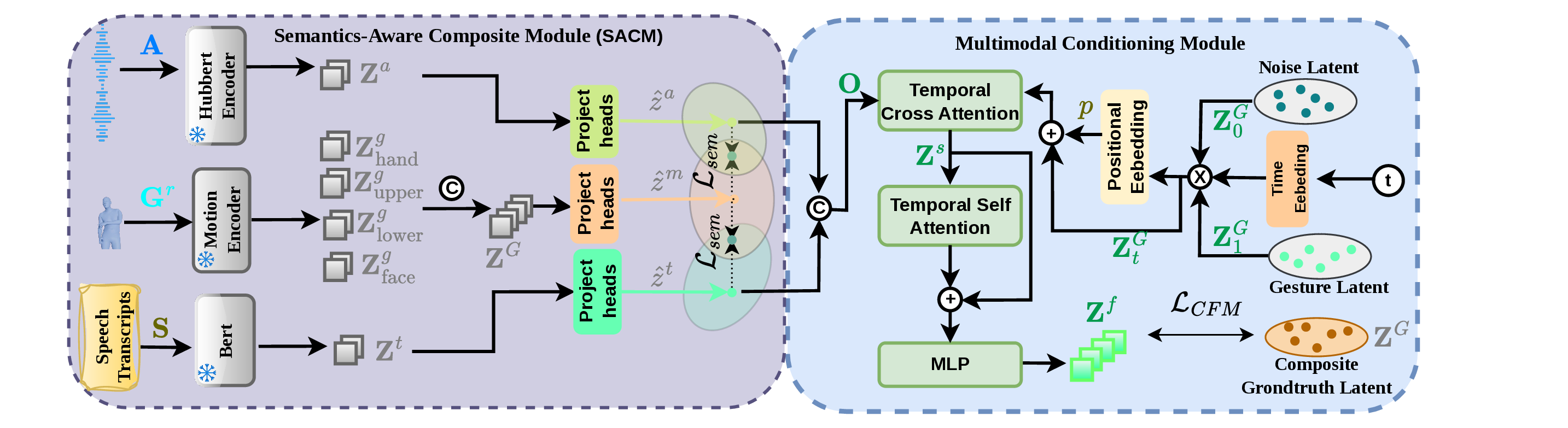}
  \caption{Our architecture comprises two synergistic modules:  (a) Semantics-Aware Composite Module (SACM), which anchors audio, textual, and motion features within an aligned semantic space to ensure cross-modal consistency; and (b) Multimodal Conditioning Module, which leverages a learned velocity field conditioned on multimodal priors and stochastic seed latents to synthesize expressive, kinematically coherent holistic gestures.}
  \label{fig:2}
\end{figure*}

\paragraph{\textbf{Semantics-Aware Composite Module (SACM).}}
SACM aligns \emph{text}, \emph{audio}, and \emph{holistic motion} in a shared semantic latent space (Fig.~\ref{fig:2}, left). Unlike part-wise alignment strategies that relate speech semantics to individual body regions in isolation~\cite{liu2024emage,zhang2025semtalk}, we adopt a \emph{body-compositional} view: we first aggregate all region latents from Stage~1 into a holistic motion representation, and then align this composite motion with language and acoustics using (1) a sequence-level cosine objective and (2) a CLIP-style contrastive objective.

Let $\mathbf{Z}^{g}_{r}\in\mathbb{R}^{L\times d_g}$ denote the Stage~1 latent sequence for region $r\in\mathcal{R}$,
where $\mathcal{R}=\{\text{hand},\text{upper},\text{lower},\text{face}\}$, $ d_g$ represent latent dimensions and $L$ is the latent temporal length.
We form a composite gesture latent by concatenation and scale normalisation:
\begin{equation}
\mathbf{Z}^{G} \;=\; \tfrac{1}{s}\,\mathrm{Concat}\!\left(
\mathbf{Z}_{\text{hand}}^{g},\mathbf{Z}_{\text{upper}}^{g},\mathbf{Z}_{\text{lower}}^{g},\mathbf{Z}_{\text{face}}^{g}
\right),
\end{equation}
where $s$ is a constant normalisation factor.
For semantics, a text encoder $\mathcal{T}$ (BERT~\cite{devlin2019bert}) maps transcripts to $\mathbf{Z}^{t}\in\mathbb{R}^{L\times d_t}$ and
an audio encoder $\mathcal{A}$ (HuBERT~\cite{hsu2021hubert}) maps waveforms to $\mathbf{Z}^{a}\in\mathbb{R}^{L\times d_a}$.
Both modalities are temporally aligned to the motion latent length $L$.
To compare heterogeneous modalities, we apply modality-specific projection heads $\Psi^{t},\Psi^{a},\Psi^{m}$ into a common $d$-dimensional space
followed by $\ell_2$ normalization:
\begin{equation}
\hat{\mathbf{Z}}^{t}=\mathrm{norm}(\Psi^{t}(\mathbf{Z}^{t})),\quad
\hat{\mathbf{Z}}^{a}=\mathrm{norm}(\Psi^{a}(\mathbf{Z}^{a})),\quad
\hat{\mathbf{Z}}^{m}=\mathrm{norm}(\Psi^{m}(\mathbf{Z}^{G})).
\end{equation}

To mitigate semantic drift between lexical and acoustic cues, we construct a fused (barycentric) target: $
\bar{\mathbf{Z}} \;=\; \mathrm{norm}\!\big(\alpha\,\hat{\mathbf{Z}}^{t} + (1-\alpha)\,\hat{\mathbf{Z}}^{a}\big),
\label{eq:fused-semantic}$
where $\alpha\in[0,1]$.

\noindent\textbf{(1) Sequence-level cosine alignment.}
Given a batch of size $B$ and time index $i\in\{1,\dots,L\}$, we minimise the mean cosine distance over time and batch:
\begin{equation}
\mathcal{L}_{\mathrm{cos}}
=
1-\frac{1}{BK}\sum_{b=1}^{B}\sum_{l=1}^{L}
\big(\bar{\mathbf{z}}_{b,l}\big)^{\!\top}\hat{\mathbf{z}}^{m}_{b,l}.
\label{eq:cos-align}
\end{equation}

\noindent\textbf{(2) CLIP-style InfoNCE alignment.}
To obtain batch-level discrimination with explicit negatives, we compute a single embedding per clip by temporal pooling
(e.g., mean pooling) $\mathbf{u}^{x}_{b}=\mathrm{Pool}(\hat{\mathbf{Z}}^{x}_{b})\in\mathbb{R}^{d}$ for $x\in\{m,a,t\}$.
The (one-way) InfoNCE loss is:
\begin{equation}
\mathcal{L}_{\mathrm{NCE}}(\mathbf{u}^{p},\mathbf{u}^{q})
=
-\frac{1}{B}\sum_{b=1}^{B}
\log
\frac{\exp((\mathbf{u}^{p}_{b})^{\!\top}\mathbf{u}^{q}_{b}/\tau)}
{\sum_{l=1}^{B}\exp((\mathbf{u}^{p}_{b})^{\!\top}\mathbf{u}^{q}_{l}/\tau)},
\end{equation}
where $\tau$ is a temperature parameter, and negatives are other samples in the mini-batch.
We use a symmetric variant:
$\mathcal{L}_{\mathrm{NCE}}^{\mathrm{sym}}(\mathbf{u}^{p},\mathbf{u}^{q})
=\tfrac{1}{2}\big(\mathcal{L}_{\mathrm{NCE}}(\mathbf{u}^{p},\mathbf{u}^{q})
+\mathcal{L}_{\mathrm{NCE}}(\mathbf{u}^{q},\mathbf{u}^{p})\big)$, and define:
\begin{equation}
\mathcal{L}_{\mathrm{clp}}
=
\tfrac{1}{2}\Big(
\mathcal{L}_{\mathrm{NCE}}^{\mathrm{sym}}(\mathbf{u}^{m},\mathbf{u}^{a})
+
\mathcal{L}_{\mathrm{NCE}}^{\mathrm{sym}}(\mathbf{u}^{m},\mathbf{u}^{t})
\Big).
\end{equation}

The overall SACM objective is: $\mathcal{L}_{\mathrm{sem}}
=\lambda_{\mathrm{cos}}\mathcal{L}_{\mathrm{cos}}
+\lambda_{\mathrm{clp}}\mathcal{L}_{\mathrm{clp}}.
\label{eq:final-semantic}$
Overall, SACM anchors holistic motion latents in a joint speech-semantic manifold, improving cross-region coherence while preserving alignment with both lexical meaning and acoustic prominence.
\paragraph{\textbf{Flow Matching Through Multimodal Conditioning.}}
Given the aligned multimodal representations, we formulate gesture generation as learning a conditional velocity (flow) field that transports samples
from a latent noise distribution to the motion manifold in the composite gesture-latent space (Fig.~\ref{fig:2}, right).
We construct a multimodal conditioning sequence by concatenating audio and text embeddings along the channel dimension
and projecting them to a unified conditioning space:
\begin{equation}
\mathbf{O}
=
\mathrm{Proj}_{O}\!\left(\mathrm{Concat}(\hat{\mathbf Z}^{a},\hat{\mathbf Z}^{t})\right)
\in\mathbb{R}^{L\times d_O}.
\end{equation}

Let $\mathbf{Z}^{G}_{1}\in\mathbb{R}^{L\times d_G}$ denote the (Stage~1) composite gesture latent of the ground-truth motion
and let $\mathbf{Z}^{G}_{0}\sim\mathcal{N}(\mathbf{0},\mathbf{I})$ denote a noise latent of the same shape.
We sample a \emph{flow time} $t\sim\mathcal{U}(0,1)$ (distinct from the text-modality superscript) and define the linear interpolant:
\begin{equation}
\mathbf{Z}^{G}_{t}
=
(1-t)\,\mathbf{Z}^{G}_{0}
+
t\,\mathbf{Z}^{G}_{1},
\label{eq:linear-interp}
\end{equation}
whose corresponding target velocity field is constant:
\begin{equation}
\hat{\mathbf v}
=
\frac{\partial \mathbf{Z}^{G}_{t}}{\partial t}
=
\mathbf{Z}^{G}_{1}-\mathbf{Z}^{G}_{0}.
\label{eq:positive_v}
\end{equation}

Before conditioning, we add a learned body-structure positional embedding $\mathbf p$ to the motion latent tokens to encode intra-frame body structure.
We then apply a temporal cross-attention block (\textit{TCAM}) to fuse the multimodal condition $\mathbf O$ with the noised motion latent:$
\mathbf{Z}^{s}=\mathrm{TCAM}\!\left(\mathbf{Z}^{G}_{t}+\mathbf p,\ \mathbf O\right)$,
where $\mathbf{Z}^{G}_{t}+\mathbf p$ provides the queries and $\mathbf O$ provides keys/values, yielding conditioned motion features
$\mathbf{Z}^{s}\in\mathbb{R}^{L\times d_s}$ (with $d_s{=}1024$ in our implementation).
Next, temporal self-attention captures global motion context to promote temporal coherence, followed by a position-wise MLP:$\mathbf{Z}^{f}
=\mathrm{MLP}\!\left(\mathrm{SelfAttn}(\mathbf{Z}^{s})\right).$
Finally, the flow network predicts the conditional velocity field $\mathbf v_{\theta}(\mathbf{Z}^{G}_{t},t\mid\mathbf O)$ from the fused features.

\begin{figure*}[t] 
    \centering \includegraphics[width=0.8\linewidth]{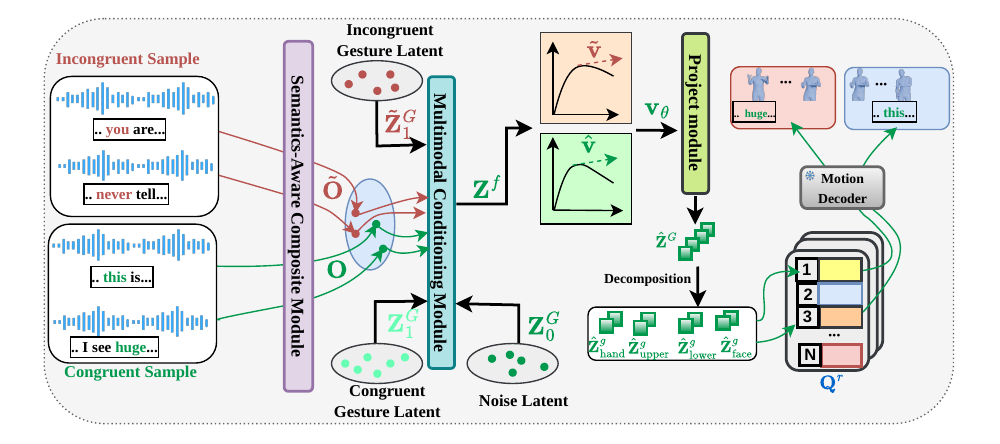} 
    \caption{Overview of the proposed semantic-aware contrastive flow matching framework for gesture generation. Audio and text transcripts are encoded into a shared semantic representation where congruent and incongruent gesture latents form contrastive pairs. Contrastive flow matching then learns motion trajectories aligned with the correct semantic context while diverging from mismatched inputs, producing coherent full-body gestures.}
    \label{fig:3} 
\end{figure*} 

\subsubsection{Contrastive Flow for Gesture Generation}
To encourage the conditional flow to remain sensitive to multimodal semantics (and reduce semantic averaging), we add a contrastive regularisation term inspired by contrastive conditional flow formulations~\cite{stoica2025contrastive}.
In Fig.~\ref{fig:3}, during training, for each sample we construct an \emph{incongruent} conditioning sequence $\tilde{\mathbf O}$
by mismatching audio--text pairs (e.g., via an in-batch random permutation of transcripts while keeping audio fixed, or vice versa).

Let $\mathbf{Z}^{G}_{1}$ denote the ground-truth composite gesture latent and $\mathbf{Z}^{G}_{0}\!\sim\!\mathcal{N}(\mathbf{0},\mathbf{I})$ the noise latent.
We sample $t\!\sim\!\mathcal{U}(0,1)$ and define the interpolated latent $\mathbf{Z}^{G}_{t}$ as in Eq.~\ref{eq:linear-interp}.
Under the straight-line path, the positive target velocity is (Eq.~\ref{eq:positive_v}).
To form a negative (incongruent) target, we sample a mismatched gesture latent $\tilde{\mathbf{Z}}^{G}_{1}$
from the example associated with $\tilde{\mathbf O}$, while reusing the same noise seed $\mathbf{Z}^{G}_{0}$ for comparability.
The corresponding negative velocity is: $
\tilde{\mathbf v} \;=\; \tilde{\mathbf{Z}}^{G}_{1}-\mathbf{Z}^{G}_{0}.$

Given the conditional velocity predictor $\mathbf v_\theta(\mathbf{Z}^{G}_{t},t \mid \mathbf O)$, we optimize the following
contrastive flow-matching objective:
\begin{equation}
\mathcal{L}_{\mathrm{CFM}}
=
\mathbb{E}\Big[
\big\|\mathbf v_\theta(\mathbf{Z}^{G}_{t},t \mid \mathbf O)-\hat{\mathbf v}\big\|_2^2
\;-\;
\lambda\,\big\|\mathbf v_\theta(\mathbf{Z}^{G}_{t},t \mid \mathbf O)-\tilde{\mathbf v}\big\|_2^2
\Big],
\qquad \lambda\in[0,1),
\label{eq:cfm}
\end{equation}
where $\lambda$ controls the strength of contrastive regularisation.
Intuitively, the first term attracts the predicted velocity toward the congruent trajectory induced by $\mathbf O$,
while the second term repels it from trajectories associated with the incongruent multimodal context $\tilde{\mathbf O}$.

\smallskip
\noindent\textbf{Sampling and decoding.}
At inference time, we sample $\mathbf{Z}^{G}_{0}\sim\mathcal{N}(\mathbf 0,\mathbf I)$ and integrate the learned conditional ODE
$\frac{d\mathbf Z}{dt}=\mathbf v_{\theta}(\mathbf Z,t\mid\mathbf O)$ to obtain a generated holistic latent
$\hat{\mathbf Z}^{G}_{1}$.
We apply a lightweight projection head to map the generated latent to a stable manifold compatible with the Stage~1 VQ codebooks:
\begin{equation}
\hat{\mathbf Z}^{G}
=
\mathrm{Proj}\!\left(\hat{\mathbf Z}^{G}_{1}\right).
\end{equation}
We then decompose the holistic latent back into region-specific latents (inverse of the concatenation order):
\begin{equation}
\hat{\mathbf Z}^{G}
=
\bigoplus_{r \in \mathcal R}
\hat{\mathbf Z}_{r}^{g},
\quad
\mathcal R = \{\text{hand}, \text{upper}, \text{lower}, \text{face}\}.
\end{equation}
Each region latent is quantised by the corresponding VQ codebook $\mathbf Q^{r}$:
\begin{equation}
\mathbf Z_{r}^{q}
=
\mathrm{Quant}\big(\hat{\mathbf Z}_{r}^{g}; \mathbf Q^{r}\big).
\end{equation}
Finally, the motion decoder $\mathcal D_{m}$ reconstructs the full-body gesture sequence:
\begin{equation}
\hat{\mathbf G}
=
\mathcal D_{m}\!\left(\left\{\mathbf Z_{r}^{q}\right\}_{r\in\mathcal R}\right).
\end{equation}

\subsubsection{Training Objective}
The comprehensive training objective for the semantic gesture generation integrates contrastive flow matching loss and semantic alignment loss:
\begin{equation}
\mathcal{L}_{\mathrm{total}} = \lambda_{\mathrm{CFM}} \mathcal{L}_{\mathrm{CFM}} + \lambda_{\mathrm{sem}} \mathcal{L}_{\mathrm{sem}}.
\end{equation}
The hyperparameters $\lambda_{\mathrm{CFM}}$ and $\lambda_{\mathrm{sem}}$ balance the contribution of optimization objective.

\section{Experiment}
\label{sec:Experiment}
\subsection{Datasets}
Our methodology is evaluated on two datasets, namely BEAT2\cite{liu2024emage} and SHOW \cite{yi2023generating}, objectively and subjectively. 
BEAT2\cite{liu2024emage} presents a large-scale multimodal motion capture dataset comprising 60 hours of synchronised speech and motion data from 25 speakers (12 female, 13 male). The dataset is organised into BEAT2-Standard (27 hours) and BEAT2-Additional (33 hours) subsets, with the latter capturing more spontaneous and natural gestural behaviours. BEAT2 significantly advances beyond its predecessor through refined mesh-based representation and MoSh++\cite{mahmood2019amass} optimisation, which enhances body proportion accuracy, finger articulation precision, and facial expression fidelity. The dataset facilitates comprehensive multimodal learning with robust cross-modal alignment between speech audio and full-body motion coded in SMPL-X parameters \cite{pavlakos2019expressive}. The BEAT2-Standard subset\cite{liu2024emage} is split with an 85\%/7.5\%/7.5\% train/validation/test division across all 25 speakers. 

SHOW dataset\cite{yi2023generating} is a comprehensive audiovisual dataset featuring 26.9 hours of synchronized speech and 3D motion data from four speakers. The dataset provides high-quality 30fps SMPL-X\cite{pavlakos2019expressive} body meshes with holistic full-body, hand, and facial expressions paired with 22kHz audio. Each video undergoes rigorous processing through PyMAF-X \cite{zhang2023pymaf} incorporating facial landmarks and shape alignment to ensure motion accuracy. The dataset emphasizes speech-gesture synchronization and rhythm correlation, segmented into 10-second clips totaling 12,019 sequences. Following established protocols\cite{yi2023generating}, the data is split into 80\%/10\%/10\% for the train/val/test division across all 4 speakers.

\subsection{Experimental Setup}

\paragraph{Implementation Details.}
All experiments are conducted on a single NVIDIA A100 GPU. For the BEAT2 dataset \cite{liu2024emage}, our training follows a two-stage pipeline. In the first stage, we train four separate body-part-specific temporal Residual Vector Quantised VAEs (RVQVAE). This stage is optimized for 800 epochs (approximately 140 hours)
with a batch size of 64 and a clip length of 64 frames. In the second stage, the semantic gesture generator is optimised for 1000 epochs (approximately 63 hours)
using the Adam optimiser with a constant learning rate of $1\times10^{-3}$, a batch size of 128, and a clip length of 64. For the SHOW dataset \cite{yi2023generating}, we extend the temporal window to a clip length of 196 frames by following the setup in \cite{yi2023generating}. The motion representation stage for SHOW is similar to the BEAT2 configuration, while the subsequent semantic generation stage is trained for 1000 epochs with a batch size of 128 and a slightly increased learning rate of $1.5\times10^{-3}$ to accommodate the expanded temporal context. 

\paragraph{Comparison Baselines.}
We conduct comprehensive comparisons against state-of-the-art holistic gesture generation methods such as Semtalk\cite{yang2025gesturehydra}, GestureLSM\cite{liu2025gesturelsm}, EMAGE\cite{liu2024emage}, Talkshow\cite{yi2023generating} and RAGGesture\cite{mughal2025retrieving}
All methods mentioned above are trained from scratch using their original code base on the BEAT2 and SHOW dataset following the train/val/test splits as explained in Section 4.1 DataSets. Other approaches such as ~\cite{liu2025semges,ao2023gesturediffuclip,zhang2024semantic} are not included as they do not focus on holistic co-speech gesture synthesis. 

\subsection{Quantitative Results}
\subsubsection{Evaluation Metrics.}
We evaluate all methods using three complementary metrics.
(1) \textbf{Fréchet Gesture Distance (FGD)}\cite{yoon2020speech} measures the distributional similarity between generated and ground-truth gestures in a learned latent feature space. Lower FGD indicates closer alignment with the real motion distribution and improved holistic realism. (2) \textbf{Beat Consistency (BC)}\cite{li2021ai} evaluates speech–motion synchronization by measuring the alignment between kinematic motion peaks and corresponding acoustic onset patterns, reflecting rhythmic and temporal consistency. (3) \textbf{Diversity}\cite{li2021audio2gestures} is calculated as the average pairwise Euclidean distance among generated samples in the test set, assessing motion variability and the stochastic richness of the generative model.
\begin{table}[t]
\centering
\scriptsize
\setlength{\tabcolsep}{4pt}
\renewcommand{\arraystretch}{0.6}
\begin{tabular}{llccc}
\toprule
Dataset & Method & FGD$\downarrow$ & BC$\uparrow$ & Diversity$\uparrow$ \\
\midrule

\multirow{5}{*}{BEAT2(25 speakers)}
& SHOW\cite{yi2023generating}              & 6.204			  & 0.652        &82  \\
& EMAGE\cite{liu2024emage}              & 5.521			    & 0.692      & 88 \\
& RAGGesture\cite{mughal2025retrieving}      & 4.446  & 0.475      &114  \\
& GestureLSM \cite{liu2025gesturelsm}      & 4.268 &  0.525 & 112    \\
& SemTalk \cite{zhang2025semtalk}           & 4.264			 &0.727   &116    \\
& \textbf{Ours}                   & \textbf{2.247} 		& \textbf{0.780}   & \textbf{120} \\

\midrule

\multirow{5}{*}{SHOW(4 speakers)}
& SHOW \cite{yi2023generating}              & 24.35			   & 0.825       & 102 \\
& EMAGE \cite{liu2024emage}              & 22.23		   & 0.823       & 106 \\
& RAGGesture \cite{mughal2025retrieving}           &  23.54		&0.812  & 94    \\
& GestureLSM \cite{liu2025gesturelsm}                    & 22.89		 & 0.809  & 102    \\
& SemTalk\cite{zhang2025semtalk}                            & 20.17		 & 0.829 & 110    \\

& \textbf{Ours}        &\textbf{18.92}    &\textbf{0.831}   &  \textbf{112}  \\
\bottomrule
\end{tabular}
\caption{Quantitative comparison with state-of-the-art methods on the BEAT2 and SHOW datasets. $\downarrow$ and $\uparrow$ indicate whether lower or higher values are better, respectively. 
Our method achieves consistently strong performance across Fréchet Gesture Distance (FGD), Beat Consistency (BC) and Diversity metrics.}
\label{tab:beat2_show_results}
\end{table}

\subsubsection{\textbf{Quantitative Comparison Evaluation  with Baselines.}}
We provide comparison with the recent state-of-the-art methods listed above, on both BEAT2 and SHOW datasets. 
It is important to note that prior works~\cite{liu2025gesturelsm,liu2024emage,zhang2025semtalk} report results on BEAT2 under single-speaker settings, leaving multi-speaker generalization underexplored. We therefore evaluate on the full 25-speaker BEAT2 benchmark. As shown in Table~\ref{tab:beat2_show_results}, our approach achieves consistently strong performance across all metrics. On BEAT2(25 speakers), our method results in the lowest FGD value, indicating closer alignment with the ground-truth motion indicating improved motion realism. 
It further achieves the highest BC score, demonstrating improved temporal coordination between speech and gesture. Our approach also produces the most diverse gesture patterns without compromising motion realism.
On the SHOW dataset (4 speakers), our model maintains better results for FGD, BC and Diversity metrics exceeding the performance of other methods. Taken together, these results suggest that our framework generalizes effectively across datasets and speaker identities, providing consistent improvements without sacrificing motion naturalness.

\paragraph{\textbf{Ablation Study.}}
Table~\ref{tab:ablation} evaluates three design choices on BEAT2 and SHOW: 
(i) removing the Semantics-Aware Composite Alignment Module (SACM), 
(ii) restricting SACM to a single modality (text-only vs.\ audio-only), and 
(iii) replacing Contrastive Flow Matching with standard Flow matching (FM).
\begin{enumerate}
    \item \emph{SACM removal.} Removing SACM leads to a clear and consistent degradation across datasets, with the most impact on Beat Consistency and FGD metrics. This confirms that explicit semantic alignment is necessary for stable speech--motion coupling.
    
    \item \emph{Modality within SACM.} Text-only or audio-only alignment provides improved results in comparison to removing SACM, but each modality emphasizes different aspects: text-only alignment improves FGD and Diversity, whereas audio-only alignment tends to preserve temporal consistency measured through BC (while still improving the FGD and Diversity metrics). This supports the view that text and audio provide complementary supervision.
    
    \item \emph{Contrastive FM vs standard FM.} Replacing contrastive flow matching with standard FM degrades the FGD metric leading to decreased motion realism. Standard FM also have lower diversity and BS values with respect to the Contrastive Flow Matching. Overall, Contrastive Flow Matching consistently improves the overall results supporting our initial claims and contributions.
\end{enumerate}

\noindent Overall, our complete model achieves the best balance on both datasets, simultaneously improving motion realism (lowest FGD), beat consistency (highest BC), and diversity.
\begin{table}[t]
\centering
\scriptsize
\setlength{\tabcolsep}{4pt}
\renewcommand{\arraystretch}{0.6}
\begin{tabular}{llccc}
\toprule
Dataset & Method & FGD$\downarrow$ & BC$\uparrow$ & Diversity$\uparrow$ \\
\midrule

\multirow{7}{*}{BEAT2}
& w/o SACM            & 4.125 & 0.592  & 102   \\
& w/ SACM(text only)    & 3.216  &0.736 & 112\\
& w/ SACM(audio only)   & 3.849 & 0.765 & 105\\
& w/ standard FM       & 4.269 & 0.776  &  111\\
& \textbf{Ours}                   & \textbf{2.247} 		& \textbf{0.780}   & \textbf{120} \\
\midrule

\multirow{7}{*}{SHOW}
& w/o SACM               & 24.28 & 0.726 & 92    \\
& w/ SACM(text only)     & 20.47 & 0.815  & 108\\
& w/ SACM(audio only)    & 22.85 &0.821  & 98\\
& w/ standard FM         & 22.86 & 0.825  & 101 \\
& \textbf{Ours}        &\textbf{18.92}    &\textbf{0.831}   &  \textbf{112}  \\

\bottomrule
\end{tabular}
\caption{Ablation study on BEAT2 and SHOW datasets. We analyze the contribution of each component, including the Multimodal Semantics-Aware Composite Module (SACM) under different modality settings, and the Contrastive Flow Matching objective compared to Standard Flow Matching.}
\label{tab:ablation}
\end{table}

\begin{figure*}[t]
  \centering
  \includegraphics[width=0.9\linewidth]{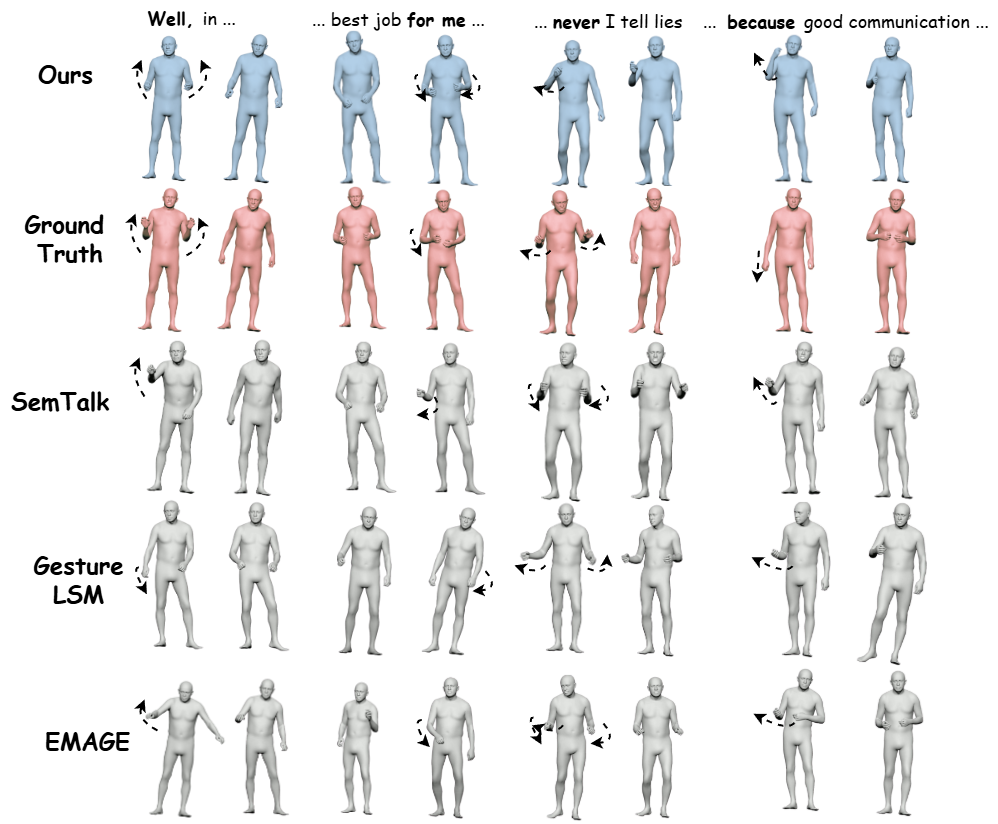}
  \caption{Qualitative comparison of semantic gesture generation on BEAT2. Our model synthesizes context-grounded gestures that precisely map to linguistic semantics: discourse marker “Well”, deictic pointing for “for me” rhythmic denial for “never” and iconic explanation for “because”. In contrast, baselines typically collapse into semantically neutral, rhythm-driven arm swings, failing to capture the distinct communicative functions of each clause.}
  \label{fig:4}
\end{figure*}
\subsection{Qualitative Results}
\subsubsection{Qualitative comparisons.}
 Figure~\ref{fig:4} compares our method against EMAGE, SemTalk, and GestureLSM on four semantically salient spans (discourse marker\cite{fraser1999discourse}, self-reference\cite{rogers1977self}, negation and causality\cite{konig2000causal}) 
 As shown in the figure, our outputs show clearer \emph{gesture stroke} realization that is consistent with the intended discourse function and the ground truth.

A discourse marker is a word or phrase that manages the flow and structure of communication, acting as a linguistic glue connecting sentences or words. For the discourse marker “Well” our model produces a synchronized beat gesture with a brief upward hand lift aligned with the stressed syllable, clearly marking discourse initiation. In contrast, SemTalk maintains rhythmic motion without emphasis-specific articulation, while EMAGE exhibits neutral or loosely synchronized movements. For the phrase “for me” our model generates a clear deictic gesture by directing the hand toward the torso during “me” forming an explicit self-referential motion aligned with lexical emphasis. SemTalk shows mild prosodic emphasis but lacks precise referential targeting, whereas EMAGE remains largely rhythm-driven. GestureLSM displays minor frame-level jitter that reduces motion coherence. For the negation phrase “never tell lies”, our model produces repeated forward index gestures synchronized with “never” reinforcing semantic negation, while baselines largely maintain neutral hand configurations. For the causal segment “because good communication” 
our framework generates a forward-expanding iconic gesture unfolding across the clause, visually supporting the explanatory structure. SemTalk shows clause-aligned motion but lacks progressive expansion, and other baselines exhibit temporal discontinuities near emphasis peaks. Overall, while SemTalk demonstrates rhythm-aligned motion with limited semantic sensitivity, our model consistently produces discourse-aware beat gestures, deictic references, and clause-level iconic expansions while maintaining smooth inter-frame transitions without noticeable jitter.

\subsubsection{User Study.}
 We recruited 30 native English speakers from the UK and US, who evaluated 24 video sequences presented in randomized order using a five-point Likert scale across three dimensions: naturalness, diversity, and alignment with speech content and timing, including ground-truth motion, EMAGE, SemTalk, and our proposed model.
\begin{figure*}[t]
  \centering
  \includegraphics[width=0.7\linewidth]{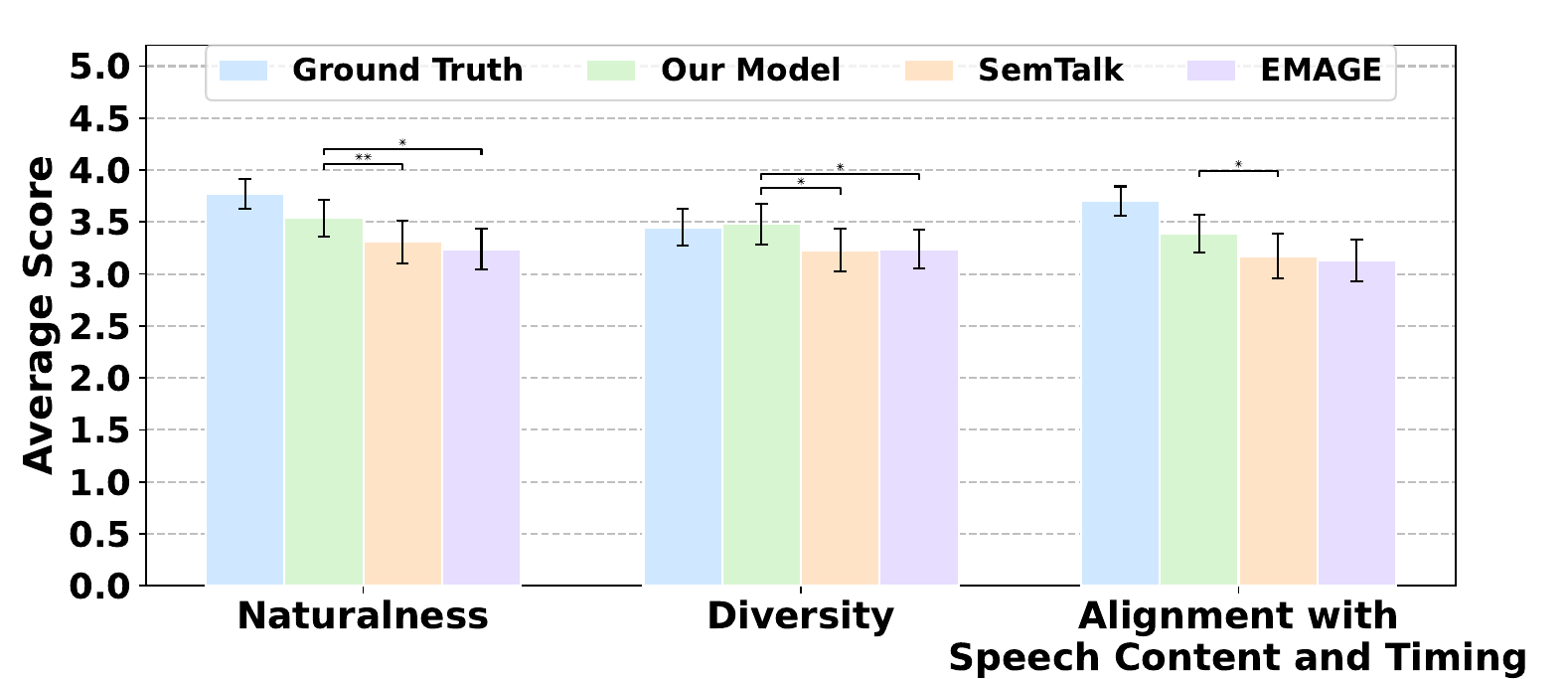}
  \caption{User study evaluation results comparing Ground Truth, Our Model, SemTalk, and EMAGE across three criteria: Naturalness, Diversity, and Alignment with Speech Content and Timing. Error bars denote standard deviation across participants. Statistical significance is indicated by \textsuperscript{*} ($p<0.05$) and \textsuperscript{**} ($p<0.01$).}
  \label{fig:5}
\end{figure*}
Details on the user study are provided in the supplementary material. 
User study results, illustrated in Fig. \ref{fig:5}, confirm that our model significantly outperforms SemTalk and EMAGE in perceptual quality. Specifically, our framework achieves higher scores in Naturalness and Alignment to Speech Content and Timing, indicating that the Semantic Coherence Module (SACM) and Contrastive Flow Matching (CFM) effectively capture nuanced speech-motion mappings. While Diversity remains similar across SemTalk and EMAGE, our approach surpasses these methods (as well the ground truth) without degrading motion realism and alignment to speech. 

\section{Conclusion}
We presented HolisticSemGes, a semantic-aware framework for holistic co-speech gesture generation that integrates speech audio, transcripts, and motion latents within a unified semantic space through the Semantics-Aware Composite Module (SACM), enabling cross-articulator coordination across the whole body. In addition, we introduce Contrastive Flow Matching-based(CFM) co-speech gesture generation model, which guides motion trajectories by aligning with semantically congruent speech–motion pairs while diverging from incongruent alternatives. Comprehensive experiments demonstrate consistent improvements in motion realism, speech–motion synchronization, semantic relevance, and gesture diversity across multiple speaker identities, producing coherent, expressive, and semantically grounded gestures. 



%
%
\bibliographystyle{splncs04}
\bibliography{main}

\clearpage
\appendix

\end{document}